\documentclass[runningheads]{llncs}

\usepackage[T1]{fontenc}
\usepackage{graphicx}
\usepackage{multirow}
\usepackage{booktabs}
\usepackage{graphicx}
\usepackage{xcolor}
\usepackage{url}

\begin{document}

\title{Evaluating Forecasting Techniques for Hardware Errors on a Large-scale HPC System}

\titlerunning{Forecasting Hardware Errors in a HPC System}

\author{
Kaiyuan Liao\inst{1} \and
Xiwei Xuan\inst{1} \and
Tanwi Mallick\inst{2} \and
Kevin Brown\inst{2} \and\\
Christopher D. Carothers\inst{3} \and
Kwan-Liu Ma\inst{1}
}

\authorrunning{K. Liao et al.}

\institute{
Department of Computer Science, University of California, Davis, USA
\email{\{kkyliao,xwxuan,klma\}@ucdavis.edu}
\and
Argonne National Laboratory, Lemont, IL, USA
\email{\{tmallick,kabrown\}@anl.gov}
\and
Rensselaer Polytechnic Institute, Troy, NY, USA
\email{chrisc@cs.rpi.edu}
}
\maketitle

\begin{abstract}

Hardware error logs in high-performance computing (HPC) systems provide early signals of abnormal behavior, yet there remain challenges in effectively forecasting these errors using modern predictive methods. This work investigates the boundaries of applying time series forecasting to HPC hardware error dynamics. We use seven years of production logs from the Theta supercomputer to evaluate the predictive efficacy of classical statistical and deep learning models. Our results show that forecasting effectiveness depends strongly on the temporal structure of the error series: regularly occurring and structurally stable errors can be modeled accurately, particularly by LSTM and Transformer architectures with temporal features, while sparse and burst-dominated errors remain difficult to predict. Rather than proposing a deployment-ready failure prediction framework, this study provides empirical guidance on when forecasting is effective and highlights potential directions for improving forecasting accuracy in HPC hardware error analysis.

\end{abstract}

%%%%%%%%%%%%%%%%%%%%%%%%%%%%%%%%%%%%%%%%%%%%%%%%%%%%%%%%
% Paper Sections
%%%%%%%%%%%%%%%%%%%%%%%%%%%%%%%%%%%%%%%%%%%%%%%%%%%%%%%%

\section{Introduction}

High-performance computing (HPC) systems, such as the Theta supercomputer previously deployed at Argonne National Laboratory, underpin large-scale scientific discovery across domains from climate modeling to drug design. Their extreme scale, thousands of compute nodes interconnected through high-speed networks and complex storage hierarchies, enables unprecedented computational capability, but also increases vulnerability to hardware faults, including memory errors, disk failures, and interconnect disruptions, which can reduce system availability and interrupt long-running workloads \cite{mohammed2017failover}. Detecting and diagnosing faults in such tightly coupled environments is challenging, as thousands of components operate concurrently under dynamic workloads, producing complex cross-layer and cascading failure dynamics \cite{bouguerra2013improving,das2018desh}. Although fault tolerance mechanisms such as checkpoint-restart are widely adopted \cite{iyer1990automatic}, they incur non-trivial storage and runtime overhead at extreme scale \cite{gainaru2012fault}. Consequently, there is growing interest in failure avoidance strategies \cite{cappello2010checkpointing}, which aim to identify early indicators of abnormal behavior \cite{gujrati2007meta} and enable timely proactive intervention.

The effectiveness of such strategies depends critically on the reliability of predictive models. Traditional statistical forecasting methods \cite{box2015time,brown2004smoothing} capture trends and periodicities but may struggle with complex, bursty temporal patterns in HPC systems \cite{siami2018comparison}. Recent advances in machine learning and deep learning have demonstrated strong forecasting capability in finance \cite{sezer2020financial}, weather \cite{hewage2020temporal}, and transportation \cite{zheng2020traffic}. However, despite abundant operational data generated by modern HPC systems, adoption of time series forecasting for HPC analysis remains limited. Existing studies have focused on predicting system metrics or performance indicators\cite{pei2023application,xu2024surrogate,tian2025hybrid}, while little attention has been paid to hardware error logs as a distinct analytical target. Hardware errors are not failures themselves but represent early signals of abnormal behavior and prerequisites for downstream fault diagnosis, prediction, or reliability assessment.

\begin{figure}[t]
\centering
\includegraphics[width=\textwidth]{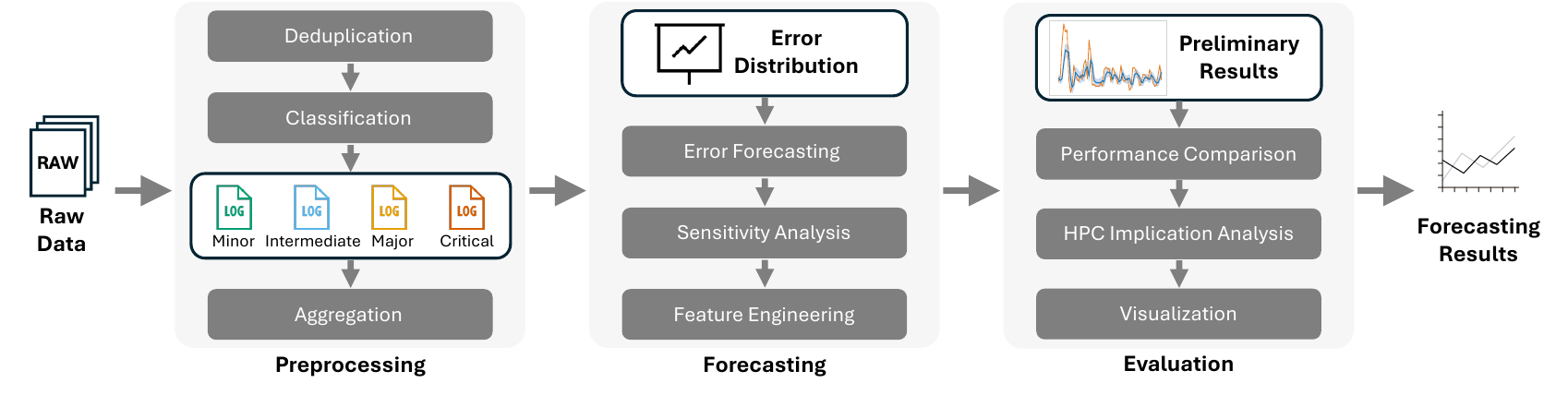}
\caption{Workflow of the HPC hardware error forecasting: 
(1) preprocessing raw logs into severity-based daily time series, 
(2) forecasting using statistical and deep learning models, and 
(3) evaluation via performance comparison and visualization.}
\label{fig:workflow}
\end{figure}

In this work, we study HPC hardware error logs from a time series forecasting perspective, aiming to determine whether and under what conditions hardware error signals exhibit predictable temporal structure. We build on prior work that introduced a severity-based classification of raw system logs, enabling their transformation into structured time series representations\cite{brown2025cug}. Leveraging this formulation, we analyze the temporal behavior of different error categories across operational regimes. Our results show that only a subset of error types demonstrates sufficient regularity for meaningful prediction, while others remain sparse and burst-dominated, limiting the effectiveness of existing forecasting methods. We benchmark statistical and deep learning models across these categories and examine the role of feature engineering, finding that LSTM and Transformer architectures capture informative temporal structure in stable series, whereas classical approaches remain competitive in certain settings. Figure~\ref{fig:workflow} summarizes our end-to-end workflow. By establishing the feasibility boundaries of forecasting in HPC error analysis, this study provides foundational insight into system error dynamics and identifies practical directions for improving forecasting accuracy under different temporal conditions.

\section{Background and Related Work} 

\subsection{HPC Log Analysis and Failure Prediction}
Large-scale HPC logs support performance characterization, anomaly detection, and predictive modeling by capturing I/O behavior, hardware events, and system diagnostics \cite{paul2020understanding,liu2020characterization,kim2023design,netti2020machine,ozer2020characterizing}. As systems scale, proactive failure prediction has become critical, leveraging anomaly detection frameworks \cite{das2020aarohi,zhong2019runtime}, supervised learning with random forests and SVMs \cite{mohammed2019failure}, and self-supervised Transformer-based lead-time prediction \cite{alharthi2023time}. While existing work emphasizes I/O profiling or discrete failure events, we focus on hardware error distributions and transform raw logs into structured time series to enable systematic forecasting-based analysis.

\subsection{Time Series Forecasting}
Time series forecasting has evolved from statistical methods like ARIMA and exponential smoothing \cite{khan2020arima,oh2024forecasting} to deep learning architectures capable of modeling complex temporal dynamics \cite{liao2026deep}. Modern models capture diverse dependencies, including CNN-based \cite{wu2023timesnet,li2025tvnet}, RNN-based \cite{matzner2025locally,kong2025unlocking}, GNN-based \cite{huang2023crossgnn,zhang2024irregular}, and Transformer-based frameworks \cite{nie2023a,wang2024card,liu2025timerxl}. Recently, LLM-inspired approaches \cite{cao2024tempo,jin2024timellm} have further advanced state-of-the-art performance. Despite this progress, applying these techniques to resilient system management remains limited. We address it by converting recurring hardware error occurrences into structured representations to evaluate their predictability in large-scale HPC systems.
\section{HPC Hardware Errors Data}

\subsection{Raw Logs Dataset}

\noindent\textbf{System Overview.}
Our study uses error logs from Theta, an 11.7-petaflop Cray XC40 supercomputer at the Argonne Leadership Computing Facility (ALCF). The system comprised 4,392 Intel Xeon Phi 7230 nodes connected via the Aries network in a 2D dragonfly topology and operated from mid-2017 to 2023. During its lifetime, Theta experienced scheduled (such as ``Big Run Mondays'') and unscheduled outages, with downtime accounting for approximately 5.33\% of operation (179 scheduled and 24 unscheduled events), most lasting less than 24 hours.

\medskip

\begingroup
\renewcommand{\thefootnote}{}
\footnotetext{\textsuperscript{*}\,Theta hardware error dataset: \url{https://reports.alcf.anl.gov/data/theta.html#THETA_HARDWARE_ERROR}}
\endgroup

\noindent\textbf{Error Logs Dataset.}
We use the \textit{Theta hardware error dataset}\textsuperscript{*}, spanning June 2017 to January 2024. Each entry records a timestamp, error code, category, and affected component. The primary categories include \textit{Correctable Memory Errors}, \textit{Machine Check Exceptions (MCEs)}, \textit{Transaction Errors}, \textit{Transient Errors}, \textit{Critical Errors}, and \textit{Informational Errors}, covering memory faults, processor exceptions, interconnect issues, and system failures. The raw logs contain duplicated and persistent records where identical failures are repeatedly logged within short intervals, introducing artificial correlation that must be removed before constructing forecasting-ready time series.

\subsection{Data Preprocessing}

\noindent\textbf{Redundancy Removal.}
We follow the cleaning procedures described in~\cite{brown2025cug}. Two forms of redundancy are present. First, exact duplicates sharing identical timestamps, components, and error codes are removed. Second, persistent errors, where the same component repeatedly generates log entries within a short interval, are consolidated using a $70$-second threshold, collapsing near-duplicate entries into single representative events. These steps reduce spurious correlations while preserving underlying failure behavior.

\medskip

\noindent\textbf{Error Classification.}
To reflect impact on system reliability, error codes are categorized into four severity levels following prior work \cite{brown2025cug}.

\begin{enumerate}
    \item \textbf{Minor errors} are correctable or warning-level faults handled automatically by the system, though frequent occurrences may indicate latent instability.
    \item \textbf{Intermediate errors} are non-persistent faults resolved by retries but may cause temporary performance degradation.
    \item \textbf{Major errors} represent persistent abnormal behaviors that often require reconfiguration or intervention.
    \item \textbf{Critical errors} are severe failures that disrupt operation and may require component replacement or system restart.
\end{enumerate}

\begin{figure}[t]
\centering
\includegraphics[width=0.85\textwidth]{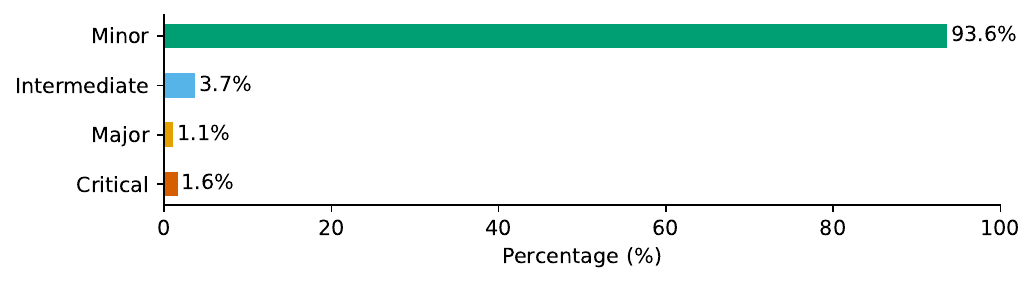}
\caption{Severity distribution of hardware errors.}
\label{fig:error_distribution}
\end{figure}

\begin{figure}[t]
\centering
\includegraphics[width=\textwidth]{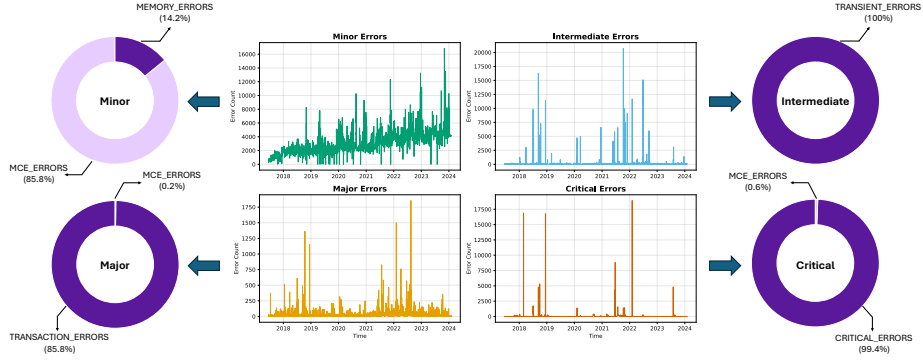}
\caption{Daily aggregated hardware error counts in Theta (2017--2024) across four severity levels with error-type composition. Minor errors show sustained activity, while Intermediate, Major, and Critical errors become sparse and spike-dominated.}
\label{fig:error_timeseries}
\end{figure}

Figure~\ref{fig:error_distribution} shows the severity distribution, where Minor errors dominate. Figure~\ref{fig:error_timeseries} presents daily aggregated counts across severity levels, illustrating temporal characteristics from stable fluctuations to sparse bursts. These differences suggest varying predictability and forecasting difficulty across error types.

\section{Model Descriptions}

We evaluate classical statistical methods and modern deep learning architectures for one-step-ahead forecasting of HPC hardware error time series. All models predict the next value \( y_{t+1} \) from a look-back window of \( L_x \) historical observations.

For deep learning models, we adopt a unified look-back window of \( L_x = 14 \), selected based on observed operational periodicity in the system. Statistical models are evaluated under multiple hyperparameter settings, and the best-performing configuration is reported. The evaluated models include:

\textbf{LAST}, a naive heuristic baseline that forecasts the next value using the most recent observation (\( y_{t+1} = y_t \)). Since it relies solely on the latest value, both the look-back and output lengths are 1.

\textbf{Rolling Mean}, a heuristic that predicts the next value as the average of recent observations. We use a look-back window of \( L_x = 3 \).

\textbf{AR} (AutoRegressive), a linear statistical model that estimates future values as weighted combinations of recent observations. We use \( L_x = 14 \).

\textbf{ARIMA} (AutoRegressive Integrated Moving Average), a classical statistical approach capable of handling non-stationary time series through differencing and autoregressive components. To improve numerical stability, we use a sliding window with \( L_x = 100 \) for re-training at each prediction step.

\textbf{HA} (Historical Average), a seasonal heuristic that predicts the next value by averaging historical observations at the same relative position within a fixed period. We set the seasonal period to \( P = 14 \).

\textbf{LSTM} (Long Short-Term Memory), a recurrent neural network architecture designed to capture long-term temporal dependencies. We have \( L_x = 14 \).

\textbf{TCN} (Temporal Convolutional Network), a convolutional model capturing temporal dependencies through causal and dilated convolutions. We use \( L_x = 14 \).

\textbf{Transformer}, an encoder-only self-attention model that learns temporal dependencies across the look-back window. We set \( L_x = 14 \).

\section{Experiments}

\subsection{Experimental Setup}

We conduct one-step-ahead forecasting on four daily time series derived from HPC system logs: \textit{Minor}, \textit{Intermediate}, \textit{Major}, and \textit{Critical} errors, each representing daily counts of a severity level. The dataset spans June 20, 2017 to January 31, 2024, totaling 2,417 observations per series, and is split chronologically (80\% training, 20\% testing). Deep learning models use a sliding window with a standardized look-back length \( L_x = 14 \), chosen to reflect the dominant biweekly maintenance cycle observed in the system.

Deep learning models are trained using Adam with mean squared error (MSE) loss, and inputs are normalized to $[0,1]$ for stability. Statistical models are evaluated via grid-searched hyperparameters with the best configuration reported. Performance is measured using MSE, MAE, and RMSE pct, and each experiment is repeated 10 times to report mean value. All experiments are conducted on NVIDIA Quadro RTX 8000 GPUs.

\begin{table}[t]
\centering
\footnotesize
\setlength{\tabcolsep}{4pt}
\renewcommand{\arraystretch}{1.05}
\caption{Forecasting performance. Bold indicates best per metric.}
\label{tab:model_perf_all}

\resizebox{\textwidth}{!}{
\begin{tabular}{lccc ccc ccc ccc}
\toprule
\textbf{Model} 
& \multicolumn{3}{c}{\textbf{Minor}} 
& \multicolumn{3}{c}{\textbf{Intermediate}} 
& \multicolumn{3}{c}{\textbf{Major}} 
& \multicolumn{3}{c}{\textbf{Critical}} \\

& MAE & MSE & R\% 
& MAE & MSE & R\% 
& MAE & MSE & R\% 
& MAE & MSE & R\% \\

\midrule
AR & 0.05430 & 0.00740 & 30.62
& 0.00569 & 0.00009 & 446.14
& 0.01484 & 0.00046 & 131.40
& 0.00357 & 0.00014 & 1282.86 \\

Rolling Mean & 0.05800 & 0.00920 & 34.18
& 0.00305 & 0.00007 & 409.54
& 0.01437 & 0.00049 & 135.57
& 0.00141 & 0.00015 & 1299.59 \\

ARIMA & 0.05410 & 0.00750 & 30.91
& 0.00357 & 0.00008 & 439.81
& 0.01519 & 0.00051 & 138.27
& 0.00236 & 0.00038 & 2094.91 \\

LAST & 0.05600 & 0.00870 & 33.31
& 0.00332 & 0.00013 & 551.36
& 0.01700 & 0.00082 & 174.90
& 0.00132 & 0.00027 & 1768.94 \\

HA & 0.12970 & 0.02740 & 59.08
& 0.00638 & 0.00011 & 495.25
& 0.01614 & 0.00050 & 136.25
& 0.00370 & 0.00016 & 1357.77 \\

\midrule
\textbf{LSTM} & \textbf{0.05397} & \textbf{0.00730} & \textbf{30.29}
& 0.00308 & \textbf{0.00007} & \textbf{405.18}
& 0.01399 & \textbf{0.00046} & 132.27
& 0.00165 & \textbf{0.00014} & \textbf{1248.98} \\

TCN & 0.05743 & 0.00817 & 32.03
& 0.00361 & 0.00008 & 435.92
& 0.01477 & 0.00051 & 140.30
& 0.00171 & 0.00015 & 1283.46 \\

Transformer & 0.05399 & 0.00777 & 31.23
& \textbf{0.00300} & \textbf{0.00007} & 411.17
& \textbf{0.01380} & 0.00047 & 134.74
& 0.00192 & \textbf{0.00014} & 1250.33 \\

\bottomrule
\end{tabular}
}

\end{table}

\subsection{Experimental Results}

\textbf{Baseline Performance on Four Error Types.}
We evaluate forecasting performance across four error types: \textit{Minor}, \textit{Intermediate}, \textit{Major}, and \textit{Critical} errors. 
Table~\ref{tab:model_perf_all} summarizes the results and illustrates how statistical and deep learning models respond to different levels of regularity and sparsity.

\begin{itemize}

\item \textit{Minor Errors.}
Deep learning models achieve the lowest overall errors on the Minor series, with LSTM consistently performing best across MAE, MSE, and RMSE\%. Statistical models such as AR and ARIMA remain competitive, indicating the presence of stable and learnable temporal dependencies that can be captured by both linear and nonlinear approaches. In contrast, heuristic baselines perform noticeably worse, suggesting that simple averaging strategies are insufficient for this moderately structured but nontrivial series. These results indicate that the Minor Error series contains meaningful temporal patterns suitable for time series forecasting.

\medskip

\item \textit{Intermediate, Major, and Critical Errors.}
For the more severe error types, performance differences across models become small and inconsistent. While certain models achieve marginal improvements in specific metrics, no method demonstrates clear dominance. These series are characterized by long periods of near-zero values interrupted by rare, high-magnitude spikes, resulting in sparse and bursty dynamics without stable temporal structure. Under such conditions, models primarily fit low-valued regions while failing to anticipate extreme peaks, and metric variations largely reflect rare events rather than genuine pattern learning. This suggests that conventional forecasting methods are inherently limited for highly irregular, event-driven fault behaviors.

\end{itemize}

\begin{figure}[t]
\centering
\includegraphics[width=0.55\textwidth]{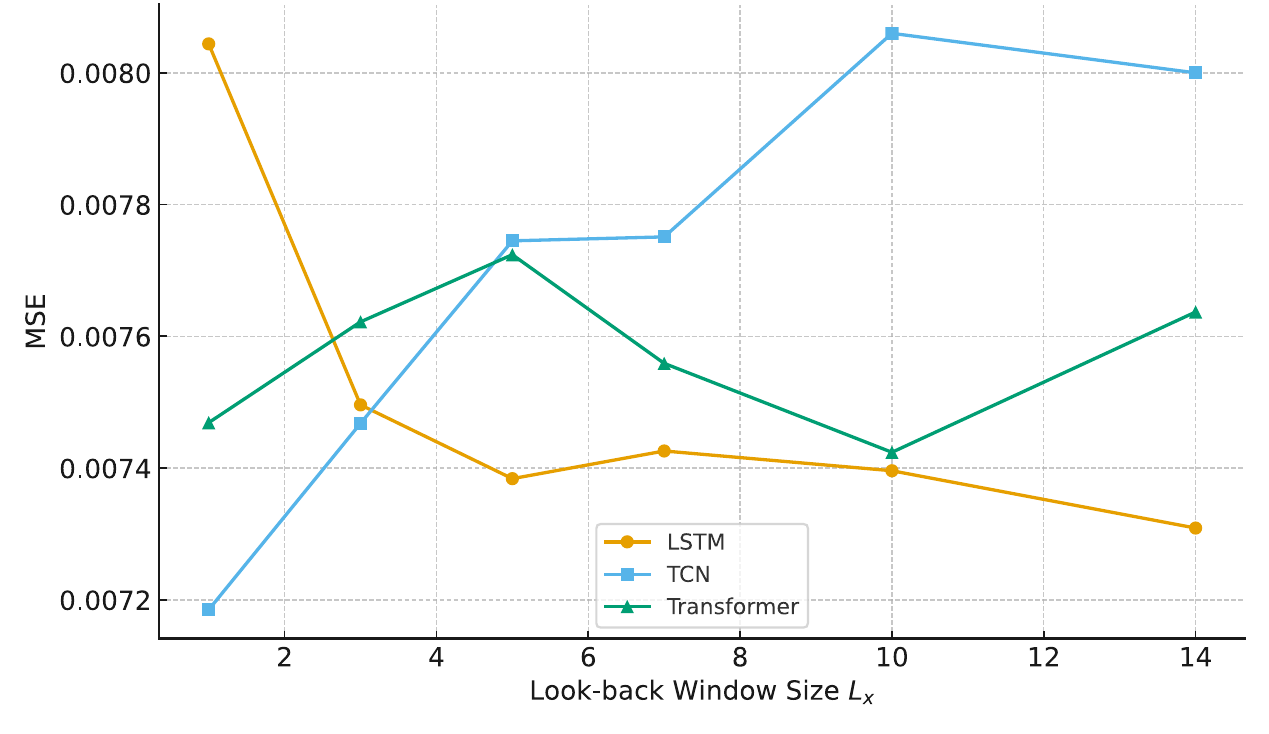}
\caption{
Impact of look-back window size $L_x$ on forecasting performance (MSE).
}
\label{fig:lookback_sensitivity}
\end{figure}

\medskip

\noindent\textbf{Sensitivity to Look-Back Window Size.}
To evaluate input length impact, we vary the look-back window \( L_x \) for each deep learning model and measure performance using MSE (Figure~\ref{fig:lookback_sensitivity}). LSTM benefits from longer context and achieves its best result at \( L_x = 14 \), whereas TCN performs best with a short window (\( L_x = 1 \)) and degrades as the window increases. Transformer remains stable across window sizes, with best performance at \( L_x = 10 \). These results suggest optimal look-back length is architecture-dependent, and model-specific tuning can improve forecasting performance beyond a unified window setting.

\medskip

\begin{table}[t]
\centering
\footnotesize
\setlength{\tabcolsep}{1pt}
\renewcommand{\arraystretch}{1.05}
\caption{Minor error forecasting performance with and without feature engineering. Bold indicates better variant.}
\label{tab:feat_eng_comparison}
\resizebox{0.7\columnwidth}{!}{
\begin{tabular}{p{2.8cm} p{2.6cm} p{2.6cm} p{2.6cm}}
\toprule
\textbf{Model} & \textbf{Metric} & \textbf{w/o FE} & \textbf{w/ FE} \\
\midrule

\multirow{3}{*}{LSTM}
 & MAE & 0.05397 & \textbf{0.05394} \\
 & MSE & 0.00730 & \textbf{0.00704} \\
 & RMSE\% & 30.29 & \textbf{29.87} \\

\midrule

\multirow{3}{*}{TCN}
 & MAE & \textbf{0.05743} & 0.05977 \\
 & MSE & \textbf{0.00817} & 0.00851 \\
 & RMSE\% & \textbf{32.03} & 32.83 \\

\midrule

\multirow{3}{*}{Transformer}
 & MAE & \textbf{0.05399} & 0.05409 \\
 & MSE & 0.00777 & \textbf{0.00728} \\
 & RMSE\% & 31.23 & \textbf{30.38} \\

\bottomrule
\end{tabular}
}
\end{table}

\noindent\textbf{Feature Engineering on Minor Error Data.}
To further improve forecasting performance on the Minor Error series, we augment the input with additional temporal features. The Minor series is the only error type exhibiting stable and recurring patterns, making it suitable for structured feature enhancement.

We incorporate four types of features:
\begin{itemize}
    \item \textbf{Day-of-Week (dow\_0 $\sim$ dow\_6)}: one-hot encoded calendar indicators to capture weekly periodicity.
    \item \textbf{First-order Difference (value\_diff)}: day-to-day changes in error counts.
    \item \textbf{Second-order Difference (second\_diff)}: changes in the rate of variation.
    \item \textbf{Exponential Moving Average (EMA)}: a smoothed representation emphasizing recent trends (span = 5).
\end{itemize}

Table~\ref{tab:feat_eng_comparison} compares LSTM, TCN, and Transformer with and without feature engineering. Feature augmentation improves LSTM and Transformer across MAE, MSE, and RMSE pct, with LSTM achieving the lowest overall errors after integration. Transformer also shows clear gains, particularly in MSE and RMSE pct, while TCN exhibits slight performance degradation. Overall, structured auxiliary inputs benefit models with adaptive memory or attention mechanisms and enhance forecasting accuracy for temporally regular error series.

\medskip

\begin{figure}[t]
\centering
\includegraphics[width=1\textwidth]{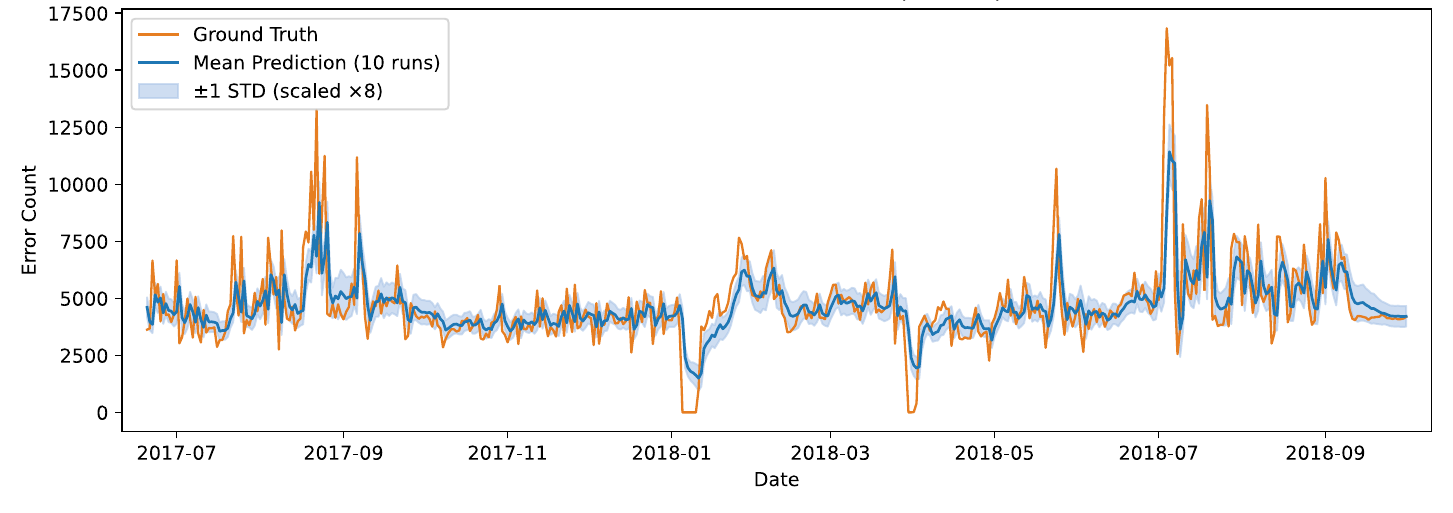}
\caption{
LSTM predictions on the Minor Error series. 
The orange line denotes ground-truth counts, and the blue line shows the mean prediction across 10 runs. 
The shaded band represents a scaled $\pm 1$ standard deviation, indicating prediction variability.
}
\label{fig:lstm_stability}
\end{figure}

\noindent\textbf{Forecast Visualization.}
Figure~\ref{fig:lstm_stability} shows LSTM forecasts on the Minor Error series, including the mean prediction and the $\pm 1$ standard deviation band across 10 independent runs. The variance remains narrow throughout the timeline (std: ± 0.00004); for clarity, the band is scaled by a factor of eight in the visualization.

\section{Discussion and Future Work}

Our results show that forecasting effectiveness is tightly coupled to the structural properties of each error series. Minor Errors exhibit recurring and gradually evolving temporal patterns and are consistently predictable using both classical and deep learning models. Performance further improves with domain-informed features, suggesting that structured auxiliary signals, like system-related indicators (e.g., day-of-week) and temporal dynamics (e.g., first-order differences), help capture multi-scale dynamics in regularly evolving system behavior.

In contrast, Intermediate, Major, and Critical Errors are sparse and burst-dominated. Across all the models and configurations we have tested, forecasting gains remain marginal, indicating that limitations arise primarily from intrinsic signal irregularity rather than model architecture. Rare and abrupt events do not provide sufficient temporal continuity to support reliable prediction within standard time series frameworks, clearly illustrating the practical boundaries of forecasting-based HPC error modeling.

Future work may extend this analysis in several directions. One important step is to evaluate generalizability across additional HPC systems with different architectures, operational policies, and workloads, which would help determine whether the observed patterns hold beyond the current dataset. Another direction is to develop clearer evaluation criteria for error-level forecasting in HPC settings. A recent survey of ML/DL-based predictive maintenance in HPC notes that accuracy expectations and standardized performance metrics for such systems remain undefined~\cite{lima2021smart}, suggesting that the usefulness threshold for hardware error forecasting is still an open question. In addition, while this study focuses on establishing foundational insights using representative deep learning models, future work can explore more recent and specialized approaches, including advanced transformer-based time series models and larger foundation models, to assess whether they provide gains in handling irregular error dynamics. Further improvements may also come from incorporating richer contextual features such as workload or topology information, as well as probabilistic or anomaly-aware modeling frameworks for better handling bursty series. Overall, forecasting is not a universal solution for HPC error analysis but a targeted tool whose effectiveness depends on temporal regularity. By identifying where forecasting succeeds and where it fails, this work highlights practical directions for improving forecasting accuracy in future HPC hardware error analysis.
\section{Conclusion}

This work evaluates the feasibility of applying time series forecasting to HPC hardware error logs using seven years of production data from the Theta supercomputer. Our results show that forecasting is effective for structurally regular error types, but remains fundamentally limited for sparse and burst-dominated series such as Intermediate, Major, and Critical Errors. We further observe that incorporating domain-informed features, such as calendar effects and short-term change indicators, improves performance for sequence models like LSTM and Transformer, but can degrade performance for other models such as TCN. These findings establish practical boundaries for forecasting-based HPC error modeling, demonstrating that its success depends on the temporal regularity of the underlying signal. Rather than proposing a deployment-ready fault prediction system, this study provides an empirical foundation for understanding when forecasting is appropriate for HPC hardware error analysis and suggests potential directions for improving forecasting accuracy through model selection, feature design, and better treatment of irregular error dynamics.

\subsubsection*{Acknowledgments} 
This work is supported in part by the U.S. Department of Energy (DOE), Office of Science via grant No. DE-SC0024580 and uses data generated at and resources of the Argonne Leadership Computing Facility, a DOE Office of Science User Facility, supported under Contract DE-AC02-06CH11357.
%%%%%%%%%%%%%%%%%%%%%%%%%%%%%%%%%%%%%%%%%%%%%%%%%%%%%%%%
% References
%%%%%%%%%%%%%%%%%%%%%%%%%%%%%%%%%%%%%%%%%%%%%%%%%%%%%%%%


\begin{thebibliography}{00}

\bibitem{mohammed2017failover}
B. Mohammed, M. Kiran, K. M. Maiyama, M. M. Kamala, and I.-U. Awan, ``Failover strategy for fault tolerance in cloud computing environment,'' Software: Practice and Experience, vol. 47, no. 9, pp. 1243--1274, 2017.

\bibitem{bouguerra2013improving}
M. S. Bouguerra, A. Gainaru, L. Bautista Gomez, F. Cappello, S. Matsuoka, and N. Maruyama, ``Improving the computing efficiency of HPC systems using a combination of proactive and preventive checkpointing,'' in 2013 IEEE 27th International Symposium on Parallel and Distributed Processing, 2013, pp. 501--512.

\bibitem{das2018desh}
A. Das, F. Mueller, C. Siegel, and A. Vishnu, ``Desh: deep learning for system health prediction of lead times to failure in HPC,'' in Proceedings of the 27th International Symposium on High-Performance Parallel and Distributed Computing, 2018, pp. 40--51.

\bibitem{iyer1990automatic}
R. K. Iyer, L. T. Young, and P. V. K. Iyer, ``Automatic recognition of intermittent failures: an experimental study of field data,'' IEEE Transactions on Computers, vol. 39, no. 4, pp. 525--537, 1990.

\bibitem{gainaru2012fault}
A. Gainaru, F. Cappello, M. Snir, and W. Kramer, ``Fault prediction under the microscope: a closer look into HPC systems,'' in SC'12: Proceedings of the International Conference on High Performance Computing, Networking, Storage and Analysis, 2012, pp. 1--11.

\bibitem{cappello2010checkpointing}
F. Cappello, H. Casanova, and Y. Robert, ``Checkpointing vs. migration for post-petascale supercomputers,'' in 2010 39th International Conference on Parallel Processing, 2010, pp. 168--177.

\bibitem{gujrati2007meta}
P. Gujrati, Y. Li, Z. Lan, R. Thakur, and J. White, ``A meta-learning failure predictor for Blue Gene/L systems,'' in 2007 International Conference on Parallel Processing (ICPP 2007), 2007, pp. 40--40.

\bibitem{box2015time}
G. E. P. Box, G. M. Jenkins, G. C. Reinsel, and G. M. Ljung, \emph{Time Series Analysis: Forecasting and Control}. John Wiley \& Sons, 2015.

\bibitem{brown2004smoothing}
R. G. Brown, \emph{Smoothing, Forecasting and Prediction of Discrete Time Series}. Courier Corporation, 2004.

\bibitem{siami2018comparison}
S. Siami-Namini, N. Tavakoli, and A. S. Namin, ``A comparison of ARIMA and LSTM in forecasting time series,'' in 2018 17th IEEE International Conference on Machine Learning and Applications (ICMLA), 2018, pp. 1394--1401.

\bibitem{sezer2020financial}
O. B. Sezer, M. U. Gudelek, and A. M. Ozbayoglu, ``Financial time series forecasting with deep learning: a systematic literature review: 2005--2019,'' Applied Soft Computing, vol. 90, p. 106181, 2020.

\bibitem{hewage2020temporal}
P. Hewage, A. Behera, M. Trovati, E. Pereira, M. Ghahremani, F. Palmieri, and Y. Liu, ``Temporal convolutional neural (TCN) network for an effective weather forecasting using time-series data from the local weather station,'' Soft Computing, vol. 24, no. 21, pp. 16453--16482, 2020.

\bibitem{zheng2020traffic}
J. Zheng and M. Huang, ``Traffic flow forecast through time series analysis based on deep learning,'' IEEE Access, vol. 8, pp. 82562--82570, 2020.

\bibitem{pei2023application}
X. Pei, M. Yuan, G. Mao, and Z. Pang, ``Application of multivariate time-series model for high performance computing (HPC) fault prediction,'' PLOS ONE, vol. 18, no. 10, p. e0281519, 2023.

\bibitem{xu2024surrogate}
X. Xu, K. A. Brown, T. Mallick, X. Wang, E. Cruz-Camacho, R. B. Ross, C. D. Carothers, Z. Lan, and K. Shu, ``Surrogate modeling for HPC application iteration times forecasting with network features,'' in Proceedings of the 38th ACM SIGSIM Conference on Principles of Advanced Discrete Simulation, 2024, pp. 93--97.

\bibitem{tian2025hybrid}
Y. Tian, X. Zhang, F. Yang, W. Yang, G. Xian, J. Yu, and L. Lai, ``A hybrid hierarchical time series model for predicting HPC job runtime,'' in 2025 10th International Conference on Computer and Communication System (ICCCS), 2025, pp. 137--142.

\bibitem{brown2025cug}
K. A. Brown, T. Mallick, Z. Lan, R. B. Ross, and C. D. Carothers, ``Analyzing a lifetime of failures on a Cray XC40 supercomputer,'' in Proceedings of the Cray User Group, 2025, pp. 103--114.

\bibitem{paul2020understanding}
A. K. Paul, O. Faaland, A. Moody, E. Gonsiorowski, K. Mohror, and A. R. Butt, ``Understanding HPC application I/O behavior using system level statistics,'' in 2020 IEEE 27th International Conference on High Performance Computing, Data, and Analytics (HiPC), 2020, pp. 202--211.

\bibitem{liu2020characterization}
Z. Liu, R. Lewis, R. Kettimuthu, K. Harms, P. Carns, N. Rao, I. Foster, and M. E. Papka, ``Characterization and identification of HPC applications at leadership computing facility,'' in Proceedings of the 34th ACM International Conference on Supercomputing, 2020, pp. 1--12.

\bibitem{kim2023design}
S. Kim, A. Sim, K. Wu, S. Byna, and Y. Son, ``Design and implementation of I/O performance prediction scheme on HPC systems through large-scale log analysis,'' Journal of Big Data, vol. 10, no. 1, p. 65, 2023.

\bibitem{netti2020machine}
A. Netti, Z. Kiziltan, O. Babaoglu, A. Sîrbu, and A. Bartolini, ``A machine learning approach to online fault classification in HPC systems,'' Future Generation Computer Systems, vol. 110, pp. 1009--1022, 2020.

\bibitem{ozer2020characterizing}
G. Ozer, A. Netti, D. Tafani, and M. Schulz, ``Characterizing HPC performance variation with monitoring and unsupervised learning,'' in International Conference on High Performance Computing, 2020, pp. 280--292.

\bibitem{das2020aarohi}
A. Das, F. Mueller, and B. Rountree, ``Aarohi: making real-time node failure prediction feasible,'' in 2020 IEEE International Parallel and Distributed Processing Symposium (IPDPS), 2020, pp. 1092--1101.

\bibitem{zhong2019runtime}
D. Zhong, A. Bouteiller, X. Luo, and G. Bosilca, ``Runtime level failure detection and propagation in HPC systems,'' in Proceedings of the 26th European MPI Users' Group Meeting, 2019, pp. 1--11.

\bibitem{mohammed2019failure}
B. Mohammed, I. Awan, H. Ugail, and M. Younas, ``Failure prediction using machine learning in a virtualised HPC system and application,'' Cluster Computing, vol. 22, no. 2, pp. 471--485, 2019.

\bibitem{alharthi2023time}
K. A. Alharthi, A. Jhumka, S. Di, L. Gui, F. Cappello, and S. McIntosh-Smith, ``Time machine: generative real-time model for failure (and lead time) prediction in HPC systems,'' in 53rd Annual IEEE/IFIP International Conference on Dependable Systems and Networks (DSN), 2023, pp. 508--521.

\bibitem{khan2020arima}
S. Khan, ``ARIMA model for accurate time series stocks forecasting,'' International Journal of Advanced Computer Science and Applications, 2020.

\bibitem{oh2024forecasting}
J. Oh and B. Seong, ``Forecasting with a combined model of ETS and ARIMA,'' Communications for Statistical Applications and Methods, vol. 31, no. 1, pp. 143--154, 2024.

\bibitem{liao2026deep}
K. Liao, X. Xuan, and K.-L. Ma, ``Deep learning for time series forecasting: A survey of recent advances,'' \textit{Frontiers of Computer Science}, vol. 20, no. 11, Art. no. 2011359, 2026.

\bibitem{wu2023timesnet}
H. Wu, T. Hu, Y. Liu, H. Zhou, J. Wang, and M. Long, ``TimesNet: temporal 2D-variation modeling for general time series analysis,'' in The Eleventh International Conference on Learning Representations, 2023.

\bibitem{li2025tvnet}
C. Li, M. Li, and R. Diao, ``TVNet: a novel time series analysis method based on dynamic convolution and 3D-variation,'' in The Thirteenth International Conference on Learning Representations, 2025.

\bibitem{matzner2025locally}
F. Matzner and F. Mráz, ``Locally connected echo state networks for time series forecasting,'' in The Thirteenth International Conference on Learning Representations, 2025.

\bibitem{kong2025unlocking}
Y. Kong, Z. Wang, Y. Nie, T. Zhou, S. Zohren, Y. Liang, P. Sun, and Q. Wen, ``Unlocking the power of LSTM for long term time series forecasting,'' in Proceedings of the AAAI Conference on Artificial Intelligence, vol. 39, no. 11, 2025, pp. 11968--11976.

\bibitem{huang2023crossgnn}
Q. Huang, L. Shen, R. Zhang, S. Ding, B. Wang, Z. Zhou, and Y. Wang, ``CrossGNN: confronting noisy multivariate time series via cross interaction refinement,'' in NeurIPS, 2023.

\bibitem{zhang2024irregular}
W. Zhang, C. Yin, H. Liu, X. Zhou, and H. Xiong, ``Irregular multivariate time series forecasting: a transformable patching graph neural networks approach,'' in Forty-first International Conference on Machine Learning, 2024.

\bibitem{nie2023a}
Y. Nie, N. H. Nguyen, P. Sinthong, and J. Kalagnanam, ``A time series is worth 64 words: long-term forecasting with transformers,'' in The Eleventh International Conference on Learning Representations, 2023.

\bibitem{wang2024card}
X. Wang, T. Zhou, Q. Wen, J. Gao, B. Ding, and R. Jin, ``CARD: channel aligned robust blend transformer for time series forecasting,'' in The Twelfth International Conference on Learning Representations, 2024.

\bibitem{liu2025timerxl}
Y. Liu, G. Qin, X. Huang, J. Wang, and M. Long, ``Timer-XL: long-context transformers for unified time series forecasting,'' in The Thirteenth International Conference on Learning Representations, 2025.

\bibitem{cao2024tempo}
D. Cao, F. Jia, S. O. Arik, T. Pfister, Y. Zheng, W. Ye, and Y. Liu, ``TEMPO: prompt-based generative pre-trained transformer for time series forecasting,'' in The Twelfth International Conference on Learning Representations, 2024.

\bibitem{jin2024timellm}
M. Jin, S. Wang, L. Ma, Z. Chu, J. Y. Zhang, X. Shi, P.-Y. Chen, Y. Liang, Y.-F. Li, S. Pan, and Q. Wen, ``Time-LLM: time series forecasting by reprogramming large language models,'' in The Twelfth International Conference on Learning Representations, 2024.

\bibitem{lima2021smart}
A. L. da Cunha Dantas Lima, V. M. Aranha, C. J. de Lima Carvalho, and E. G. Sperandio Nascimento, ``Smart predictive maintenance for high-performance computing systems: a literature review,'' The Journal of Supercomputing, vol. 77, no. 11, pp. 13494--13513, 2021.

\end{thebibliography}
\end{document}